\pdfoutput=1

\documentclass[11pt]{article}

\usepackage{EMNLP2023}

\usepackage{times}
\usepackage{latexsym}
\usepackage{placeins}
\usepackage{enumitem}
\usepackage{listings}

\usepackage[T1]{fontenc}

\usepackage[utf8]{inputenc}

\usepackage{microtype}
\usepackage{amsmath}
\usepackage{tabularx}
\usepackage{makecell}
\usepackage{graphicx}
\usepackage{tcolorbox}

\usepackage{inconsolata}
\usepackage{booktabs}
\usepackage{cleveref}
\usepackage{multirow}

\title{Multi-Turn Reasoning When Context Arrives in Pieces: Scalable Sharding and Memory-Augmented RL}

\author{
  Shu Tong Luo, Wenqin Liu, Rui Liu \\
  The University of Melbourne \\
\texttt{\{shutong2,wenqinl,ruiliu2\}@student.unimelb.edu.au} 
  \AND
  Mingming Gong \\
  The University of Melbourne \\
  \texttt{mingming.gong@unimelb.edu.au} \\\And
  Jiaxian Guo \\
  The University of Tokyo \\
  \texttt{jiaxian.guo@weblab.t.u-tokyo.ac.jp} \\
}

\begin{document}
\maketitle
\begin{abstract}


When a user reveals task-critical information across several conversation turns, LLM accuracy drops by up to 65\% despite full context availability. We show that this \textit{Lost in Conversation} degradation can be substantially mitigated by training models to maintain a compact rolling memory instead of attending to a growing history. To make such training scalable, we introduce a low-cost sharding pipeline that converts single-turn QA datasets into multi-turn fragmented-information episodes, eliminating the need for hours of manual annotation. Training only on sharded GSM8K, our memory-augmented policy significantly improves multi-turn accuracy and generalises zero-shot to harder math and out-of-domain long-context QA. Moreover, memory-trained models outperform full-history baselines even when given the full history at test time, suggesting that learning to compress induces more robust incremental reasoning than full-context exposure alone.


\end{abstract}

\section{Introduction}

\begin{figure*}[t]
    \centering
    \includegraphics[width=\textwidth]{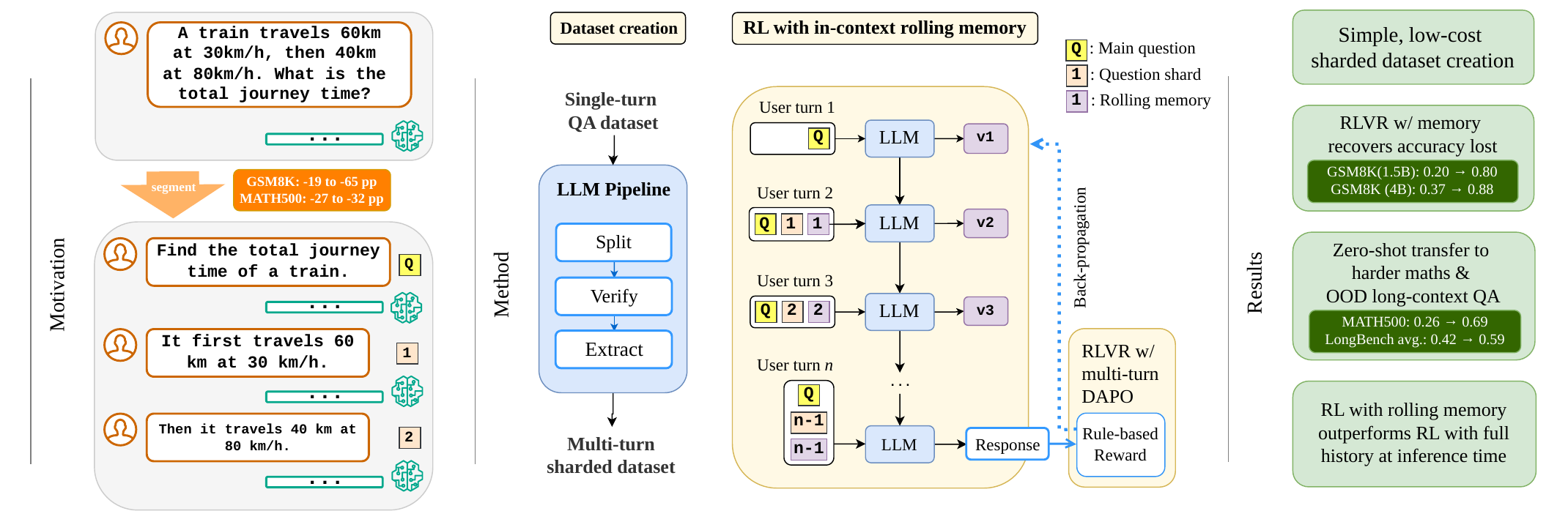}
    \caption{Overview of our approach. We shard single-turn QA problems into sharded multi-turn episodes and use RLVR to train a memory-augmented policy via multi-turn DAPO. Models trained exclusively on sharded GSM8K generalise well to in-domain and OOD long-context datasets.}
    \label{fig:overview}
\end{figure*}

Consider a user interacting with a travel-planning assistant. Rather than providing all requirements upfront, the user incrementally reveals constraints such as including the destination, budget, and travel dates across multiple turns. Successfully completing such task requires the model to continuously accumulate and reconcile information distributed throughout the conversation. Yet, frontier LLMs struggle substantially in such settings, with average task accuracy degrading by 39\% on the benchmark of \citet{lost-in-conversation}. This phenomenon is termed \textit{Lost in Conversation} (LiC). Since incremental context is fundamental to both human-AI dialogue and agentic workflows, LiC represents a major bottleneck for real-world LLM applications.

Reinforcement Learning with Verifiable Rewards (RLVR) has driven recent progress on reasoning~\citep{deepseekr1, dapo}, but two obstacles block its direct application to LiC. First, training data is expensive: the LiC benchmark pipeline requires 1 to 4 hours of manual inspection per task and 30 LLM simulations per instruction, yielding only 90 to 120 validated examples~\citep{lost-in-conversation}, insufficient for RLVR at scale. Second, models default to reasoning over the full chat history, and this strategy degrades with length~\citep{liu2023lostinthemiddle}, leaving no obvious mechanism for robust cross-turn accumulation.

We address both obstacles with a single recipe. The model maintains a bounded natural-language memory buffer that it rewrites at every turn, trained via RLVR on cheaply sharded multi-turn data. Only the buffer, not the raw history, is carried forward. While such memory buffers have been shown to compress single-turn long documents effectively~\citep{memagent}, whether the same mechanism extends to incremental dialogue, where information is partial at every step and must be integrated under reward, remains unknown. We answer this question affirmatively, and find that bounded memory is not merely a viable substitute for full history but a stronger training signal. 

Our contributions are threefold: 
(1) A cheap, scalable sharding pipeline that converts any single-turn QA dataset into multi-turn fragmented episodes using only 1 to 3 few-shot examples, with noise-augmented variants for ablation; (2) A memory-augmented, multi-turn RL recipe that recovers up to 60 points of LiC degradation on GSM8K~\cite{gsm8k} and surpasses full-history training under all evaluation conditions; (3) Evidence that this capability is domain-general: models trained only on GSM8K generalise zero-shot to MATH500~\cite{hendrycksmath2021, math500} and LongBench~\cite{longbench}.


\section{Related Work}

\paragraph{The Lost in Conversation problem.} While most multi-turn benchmarks treat conversations episodically \citep{mt-bench, mt-eval}, \citet{lost-in-conversation} show that frontier model performance degrades by 39\% on average under incremental context revelation, a setting that remains largely underexplored for training. Concurrent work \citep{rlaar} addresses solvability detection from incomplete shard sequences via RLVR, yet scalable data construction and accumulation reliability across complete sequences remain open.

\paragraph{Memory-augmented language models.} As context grows, model performance degrades due to the lost-in-the-middle phenomenon~\citep{liu2023lostinthemiddle}, motivating memory mechanisms that avoid relying on the full history. Retrieval-based approaches~\citep{rag, rag-survey} address this but require explicit retrieval infrastructure; an appealing alternative is an in-context memory buffer that the model itself maintains.~\citet{memagent} demonstrates the viability by training a model to iteratively compress document chunks into a bounded memory buffer, outperforming full-context approaches on long-context QA.

\paragraph{RLVR.}
RLVR has emerged as a powerful training paradigm for eliciting reasoning capability in LLMs \citep{deepseekr1, wen2025reinforcementlearningverifiablerewards}, with subsequent work focusing on stabilising policy optimisation~\citep{dapo, vapo, drgrpo}. MemAgent \citep{memagent} extends this to sequences of context-independent model calls, propagating a single outcome reward across all intermediate memory update steps, but operates over document chunks and has not been extended to genuinely multi-turn dialogue, where information is partial and must be integrated incrementally.

\section{Methodology}
\subsection{Sharded Dataset Construction}

We construct multi-turn sharded datasets from GSM8K and MATH500 via LLM-based sharding, which preserves the logical structure of each problem by decomposing it into semantically complete units. Each problem is processed through a three-step prompting pipeline: (1) the problem statement is segmented into its minimal logical units; (2) the segmentation is then verified for completeness and non-redundancy; (3) the core question is extracted and withheld from the shard sequence, instead presented at every turn alongside the current shard. The pipeline requires only 1--3 manually annotated few-shot examples as reference in the prompt, which significantly reduces human effort.

For long-context datasets, which lack the discrete reasoning structure that motivates LLM-based segmentation, the LongBench QA documents are instead chunked into fixed-size token segments aligned to sentence boundaries, averaging 10 to 14 shards per document.

To ablate what training structure best induces memory extraction, we train three dataset variants: (1) \textit{Sharded}, following the pipeline above; (2) \textit{Full-question}, where one randomly-positioned shard contains the full question and remaining shards contain randomly sampled Wikitext passages~\cite{wikitext}, forcing the model to distinguish relevant content from noise; and (3) \textit{Mixed}, interleaving \textit{Sharded} with Wikitext noise shards to combine logical decomposition with distraction. We additionally construct noise-augmented sets from MATH500 by injecting $n \in \{2, 4\}$ noise shards between each turn to assess test-time robustness.

\subsection{RL with In-Context Memory}

Training a model to accumulate information across fragmented turns introduces a credit assignment problem: the policy must learn to write useful memory states at every intermediate turn, yet the only training signal is whether the final answer is correct. We extend the multi-conv DAPO framework \citep{memagent} from single-turn native inference to a genuinely multi-turn setting, applying it to discrete reasoning shards derived from single-turn problems rather than chunks of a single long document.

\paragraph{Memory Mechanism.} At each shard turn $t \in \{1, \dots, K\}$, the policy receives the question $q$, the current memory state $m_t$, and the incoming shard $s_t$, and generates a rewritten memory $m_{t+1} \sim \pi_\theta(\cdot \mid q, m_t, s_t)$. The memory is a free-form natural language buffer capped at $L_m = 256$ tokens, deliberately constrained to prevent the model from copying all previously seen evidence and to make selective retention non-trivial. At the final turn, the policy conditions only on $q$ and the terminal memory $m_K$ to produce an answer $a \sim \pi_\theta(\cdot \mid q, m_K)$. The memory representation receives no direct supervision; its structure emerges from the downstream reward signal alone.

\paragraph{Training Objective.} For each sample, $G$ trajectories are sampled from 
$\pi_{\theta_\text{old}}$, each yielding a scalar reward $R^{(g)} = \mathcal{V}(a^{(g)}, 
y^\star)$ from a rule-based verifier. Group-relative advantages are computed as:
\begin{equation}
    \hat{A}^{(g)} = R^{(g)} - \frac{1}{G}\sum_{h=1}^{G} R^{(h)}.
\end{equation}
This advantage is propagated uniformly to all turns in trajectory $g$. As determining what a memory state \textit{should} contain at any given turn is non-deterministic, we avoid intermediate credit assignment. Following DAPO \citep{dapo}, the actor is updated with a dual-clipped surrogate with token-mean aggregation across all turns, ensuring longer memory updates do not disproportionately dominate the gradient relative to the final answer turn. A KL penalty is applied as a separate loss term to constrain policy drift:
\begin{equation}
    \mathcal{L}(\theta) = \mathcal{L}_\text{clip}(\theta) + 
    \beta\, D_\text{KL}(\pi_\theta \| \pi_\text{ref}).
\end{equation}

\paragraph{Full-History Baseline.} We train a separate \textit{full-history} model where the policy receives the concatenated chat history of all previous turns, rather than a compressed memory state. At inference, all models are evaluated under both the \textit{memory} and \textit{full-history} condition, isolating whether performance differences are driven by memory compression or the underlying shard structure.

\section{Experimental Setup}

We train two model scales, Qwen2.5-Math-1.5B-Instruct~\cite{yang2024qwen25mathtechnicalreportmathematical} and Qwen3-4B-Thinking~\cite{yang2025qwen3technicalreport}, with $G$=4 trajectories per prompt on 2×A100 GPUs using veRL+vLLM~\cite{verl, vllm}. The actor uses a dual-clipped PPO surrogate~\cite{ppo, dapo}, DrGRPO-style advantage normalisation~\cite{drgrpo, deepseekmath}, KL penalty ($\beta=0.001$), and learning rate $1\times10^{-6}$. 

We evaluate under two inference conditions: \textit{memory}, where the model receives only the question, current memory state, and incoming shard at each turn, and \textit{full-history}, where all previous shards are provided as context. OOD evaluation covers five LongBench subsets: 2WikiMQA~\cite{2wikimqa}, HotpotQA~\cite{hotpotqa}, MultifieldQA~\citep{longbench}, Qasper~\cite{qasper}, and TriviaQA~\cite{triviaqa}. Training is conducted exclusively on GSM8K, evaluating on the other datasets. For brevity, variant comparisons and LongBench results are 4B-only, where the larger model provides a cleaner signal.

\section{Results}

\begin{figure*}[!t]
    \centering
    \includegraphics[width=\textwidth]{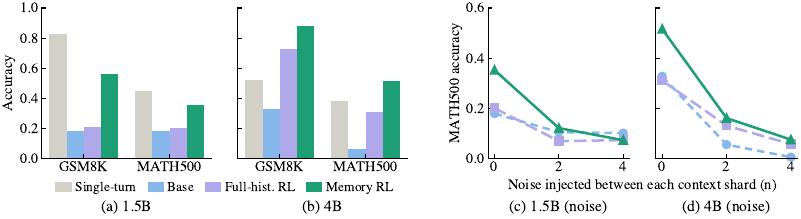}
    \caption{Accuracy under full-history inference across datasets and model scales (a--b), and MATH500 noise robustness under shard injection (c--d), also evaluated with full conversation history available. Memory RL consistently outperforms both baselines, recovering LiC degradation and maintaining an advantage under noise.}
    \label{fig:raw_eval}
\end{figure*}

\begin{table*}[!t]
\centering
\small
\begin{tabular}{llcccc}
\toprule
Model & Method & GSM8K & MATH500 & Noisy ($n$=2) & Noisy ($n$=4) \\
\midrule
\multirow{2}{*}{Qwen2.5-1.5B}
    & Base                      & 0.199 & 0.334 & 0.290 & 0.284 \\
    & Memory RL (Sharded)       & \textbf{0.799} & \textbf{0.550} 
                                & \textbf{0.290} & \textbf{0.290} \\
\midrule
\multirow{2}{*}{Qwen3-4B ($r$=512)}
    & Base                      & 0.265 & 0.050 & 0.064 & \textbf{0.050} \\
    & Memory RL (Sharded)       & \textbf{0.825} & \textbf{0.516} 
                                & \textbf{0.160} & 0.048 \\
\midrule
\multirow{5}{*}{Qwen3-4B ($r$=1024)}& Base                      & 0.370 & 0.260 & 0.232 & 0.196 \\
    & Memory RL (Sharded)       & \textbf{0.877} & 0.638 & 0.538 & 0.438 \\
    & Memory RL (Full-Question) & 0.823 & 0.682 & \textbf{0.638} & 0.516 \\
    & Memory RL (Mixed)      & 0.849 & \textbf{0.692} & 0.630 & \textbf{0.570} \\
\bottomrule
\end{tabular}
\caption{Accuracy under the rolling-memory framework. Memory RL models consistently outperform base models, generalising to unseen harder datasets. Truncation at $r{=}512$ collapses thinking model generalisation.}
\vspace{-1em}
\label{tab:main_sharded}
\end{table*}

\paragraph{Memory RL vs.\ Full-History Training}

\Cref{fig:raw_eval} reveals a substantial \textit{LiC} effect between single-turn and sharded multi-turn base models (19.4--64.9pp drop). Full-history RL partially recovers this, but Memory RL outperforms it by 15.2--35.6pp across both datasets and model scales, gains that persist under identical full-history conditions without the memory mechanism, suggesting the memory training objective instils more robust representations rather than merely exploiting compression at inference time. Injecting noise shards further exposes this gap; Full-history RL degrades sharply while Memory RL remains more resilient at $n{=}2$, consistent with selective retention implicitly encouraging noise-filtering.

\paragraph{Memory-Augmented Evaluation}

\Cref{tab:main_sharded} shows that these gains extend to the rolling-memory setting, where Memory RL produces consistent improvements across both scales, with the 1.5B and 4B ($r=1024$) models improving by 60\% and 51\% on GSM8K respectively, and both generalising to the unseen, harder MATH500 dataset, supporting that training induces general incremental reasoning rather than dataset-specific pattern matching.

The $r=512$ ablation reveals that the 4B thinking model requires sufficient response length to externalise its reasoning chain, with truncation collapsing MATH500 accuracy to 0.050 and suppressing noise robustness; both recover at $r=1024$. Among these variants, \textit{Sharded} achieves the highest in-distribution accuracy but degrades most under noise, while \textit{Mixed} offers the best overall balance, as interleaving context shards with noise during training encourages the model to both decompose problems logically and filter irrelevant content, improving resilience when noisy shards are encountered at inference time.

\paragraph{Generalisation to Long-Context QA}

\Cref{tab:cross_task_full} shows that all Memory RL variants generalise to every LongBench QA subset without task-specific training, improving average F1 by 31.7--43.4\% over the base model. \textit{Full-Question} and \textit{Mixed} outperform \textit{Sharded} on every subset, suggesting that training on noisy contexts builds distractor-filtering skills that transfer more readily to open-domain QA than \textit{Sharded}'s sequential reasoning. Overall, these results indicate that memory-based RL induces a general-purpose information extraction capability that transfers broadly across domains.

\begin{table}[t]
\centering
\small
\begin{tabular}{lccccc}
\toprule
Model & 2Wiki & Hotpot & Multi & Qasper & Trivia \\
\midrule
Base                 & 0.470 & 0.407 & 0.260 & 0.256 & 0.691 \\
Sharded           & 0.682 & 0.588 & 0.392 & 0.307 & 0.775 \\
Full-Q     & 0.705 & 0.616 & \textbf{0.413} & 0.375 & 0.850 \\
Mixed             & \textbf{0.713}& \textbf{0.627}& 0.406& \textbf{0.390} & \textbf{0.852}\\
\bottomrule
\end{tabular}
\caption{LongBench F1 score for training variants (4B)}
\label{tab:cross_task_full}
\vspace{-1em}
\end{table}

\section{Conclusion}

We presented a scalable sharding pipeline for constructing multi-turn training episodes from single-turn QA datasets, and showed that training a memory-augmented policy via multi-turn DAPO substantially recovers \textit{LiC} degradation. Memory-trained models outperform full-context baselines even under full-context evaluation, and generalise zero-shot to harder mathematical reasoning and out-of-domain long-context QA, suggesting that memory compression induces a domain-agnostic incremental reasoning capability. We aim for this work to provide a practical and reproducible foundation for future research on LLM reliability.

\section*{Limitations}



All datasets and models used are publicly available for research purposes. GSM8K and both Qwen models are released under permissive open-source licenses (MIT and Apache 2.0 respectively); sharded derivatives produced in this work are likewise intended for research use only.

Due to resource constraints, experiments are conducted on two scales within a single model family (Qwen). While the lost-in-conversation phenomenon has been demonstrated consistently across all tested model families \citep{lost-in-conversation}, and we therefore expect our findings to generalise, cross-family validation remains future work. 

The memory buffer is intentionally capped at 256 tokens, stricter than the shard content itself, to force the model to learn selective retention rather than verbatim copying; larger buffer sizes may relax this inductive bias at
the cost of less compact representations. 

Finally, while our models generalise zero-shot to LongBench QA subsets, these are evaluated without task-specific training, meaning performance may not reflect the full potential of the memory mechanism under domain-matched conditions.

\bibliography{anthology,custom}
\bibliographystyle{acl_natbib}

\appendix

\section{Appendix}
\label{sec:appendix}


\subsection{Training Hyperparameters}

All models are trained for 10 epochs on sharded GSM8K with a batch
size of 64 on 2$\times$A100 GPUs, taking an average of approximately 35 hours per run until convergence. Sequence lengths are deliberately constrained during training. The memory buffer is capped at 256 tokens to force selective retention rather than verbatim copying, and the maximum number of shards per episode is capped at 35 to bound trajectory length. 

\begin{table}[h]
\centering
\begin{tabular}{ll}
\toprule
\textbf{Parameter} & \textbf{Value} \\
\midrule
Training epochs & 10 \\
Train batch size & 64 \\
Max shard tokens & 1,024 tokens \\
Max memory buffer $L_m$ & 256 tokens \\
Max shards per episode & 35 \\
Max response length & 512 / 1,024 tokens \\
LR warmup steps & 10 \\
Wall-clock time & $\sim$30 hours / run\\
\bottomrule
\end{tabular}
\caption{RL training configuration. Max response length 512/1,024
corresponds to 1.5B/4B models respectively.}
\label{tab:impl}
\end{table}

The response length budget differs by model scale:
the 1.5B model is a math-specialised model and produces compact
answers, so $r{=}512$ suffices. The 4B model is a general-purpose
thinking model that expends a substantial portion of its response
budget on chain-of-thought reasoning before producing an answer;
at $r{=}512$ this reasoning is truncated, collapsing MATH500 accuracy
to 0.050. Setting $r{=}1024$ restores meaningful performance, as
reported in \Cref{tab:main_sharded}. 
 
\subsection{Sharding Prompt}

\begin{figure}[!htbp]
\centering
\begin{tcolorbox}[title=Sharding Prompt,
                  colback=gray!10, colframe=black]
\footnotesize\ttfamily
You are given a math problem. Rewrite it as a natural multi-turn
conversation. Each turn should restate part of the original problem
in conversational form, preserving ALL information exactly.\\

\textbf{Rules:} \\
1. You may rephrase, but preserve all numbers, expressions, symbols, constraints, and formatting. \\
2. Only describe what is explicitly stated. Do not add reasoning, methods, hints, or interpretations.\\
3. Do not add or remove facts. \\
4. Output only JSON with key \texttt{shards}. \\

\textbf{Example 1:}\\
\textit{Q: Betty is saving money for a new wallet which costs \$100. Betty has only half of the money she needs. Her parents decided to give her \$15 for that purpose, and her grandparents twice as much as her parents. How much more money does Betty need to buy the wallet?} 

\begin{tcolorbox}[colback=white, colframe=gray!30,
                  left=2pt, right=2pt, top=1pt, bottom=1pt,
                  boxsep=0pt]
\{"shards": [\{"shard": "How much more money does Betty need to buy the wallet?"\},
    \{"shard": "The wallet costs \$100"\},
    \{"shard": "Betty has only half of the money she needs"\},
    \{"shard": "Her parents gave her \$15"\},
    \{"shard": "Her grandparents gave twice as much as her parents"\}]\}
\end{tcolorbox}

\textbf{Example 2:}\\
\textit{Q: What is the value of $3^4 \times 3^2$?} 

\begin{tcolorbox}[colback=white, colframe=gray!30,
                  left=2pt, right=2pt, top=1pt, bottom=1pt,
                  boxsep=0pt]
\{"shards": [\{"shard": "An expression is given: $3^4 \times 3^2$."\},\\
\{"shard": "The question asks for its value."\}]\}
\end{tcolorbox}

\textbf{Example 3:}\\
\textit{Q: Convert the point $(0,3)$ in rectangular coordinates to
polar coordinates. Enter your answer in the form $(r, \theta)$,
where $r > 0$ and $0 \leq \theta < 2\pi$.}

\begin{tcolorbox}[colback=white, colframe=gray!30,
                  left=2pt, right=2pt, top=1pt, bottom=1pt,
                  boxsep=0pt]
\{"shards": [\{"shard": "A point is given: $(0, 3)$ in
rectangular coordinates."\}, \{"shard": "It should be converted to
polar coordinates."\}, \{"shard": "The answer format is $(r, \theta)$,
where $r > 0$ and $0 \leq \theta < 2\pi$."\}]\}
\end{tcolorbox}

\medskip
\textit{Q: \texttt{[[QUESTION]]}}
\end{tcolorbox}
\caption{Sharding prompt used to construct multi-turn episodes from
single-turn problems. \texttt{[[QUESTION]]} is replaced with the
problem text at runtime.}
\label{fig:shard-prompt}
\end{figure}
The segmentation prompt is shown in \Cref{fig:shard-prompt}, representative of the validation and question extraction prompts which follow the same format. It includes explicit rules instructing the model to retain all numerical and symbolic content verbatim while avoiding added reasoning, hints, or interpretations, with output constrained to JSON for reliable programmatic parsing. Three few-shot examples spanning simple arithmetic, symbolic expressions, and multi-constraint geometry encourage generalisation across diverse problem types, and the approach can readily extend to other domains by substituting the few-shot examples accordingly.

\FloatBarrier

\subsection{Memory Mechanism}

Prompt templates for intermediate memory update turns and the final
answer turn are shown in \Cref{fig:memory-prompt}. The final answer turn requires the model to produce its answer in \verb|\boxed{}| format, enabling rule-based match verification against the ground-truth solution without any learned reward model.

\begin{figure}[!htbp]
\centering
\begin{tcolorbox}[title=Intermediate Turn,
                  colback=gray!10, colframe=black,
                  left=2pt, right=2pt, top=2pt, bottom=2pt]
{\footnotesize\ttfamily\raggedright
You are given a question and additional information. Please read the
new information carefully and update your understanding to help answer
the question. Retain all relevant details from your previous
understanding while incorporating the new information.\\[4pt]
\textlangle question\textrangle\ \{question\}\ \textlangle/question\textrangle\\
\textlangle memory\textrangle\ \{memory\}\ \textlangle/memory\textrangle\\
\textlangle segment\textrangle\ \{segment\}\ \textlangle/segment\textrangle\\[4pt]
Updated memory:}
\end{tcolorbox}
\begin{tcolorbox}[title=Final Turn,
                  colback=gray!10, colframe=black,
                  left=2pt, right=2pt, top=2pt, bottom=2pt]
{\footnotesize\ttfamily\raggedright
You are presented with a problem and a previous memory. Please answer
the problem based on the previous memory and put the answer in
\textbackslash boxed\{\}.\\[4pt]
\textlangle question\textrangle\ \{question\}\ \textlangle/question\textrangle\\
\textlangle memory\textrangle\ \{memory\}\ \textlangle/memory\textrangle\\[4pt]
Your final answer:}
\end{tcolorbox}
\caption{Prompt templates for intermediate memory update turns and
the final answer turn. At intermediate turns the model rewrites its
memory given a new shard; at the final turn it answers using only
the terminal memory state, with no access to the shard history.}
\label{fig:memory-prompt}
\end{figure}

\end{document}